\documentclass[10pt,conference,letterpaper]{IEEEtran}
\IEEEoverridecommandlockouts
\usepackage{ragged2e}
\usepackage{cite}
\usepackage{textcomp}
\usepackage{xcolor}
\usepackage{graphicx}
\usepackage{url}
\usepackage{xspace}
\usepackage{latexsym}
\usepackage{algorithm}
\usepackage{algorithmic}
\usepackage{paralist}
\usepackage{amsmath}
\usepackage{amssymb}
\usepackage{amsfonts}
\usepackage{multirow}
\usepackage{subfigure}
\usepackage{booktabs}
\usepackage{color}
\usepackage{threeparttable}

\newtheorem{example}{Example}

\newcommand{\authorcop}{$^{\ast}$}
\newcommand{\authorsep}{\hspace{5mm}}
\newcommand{\authorone}{$^{\dagger}$}
\newcommand{\authortwo}{$^{\ddagger}$}
\newcommand{\authorthree}{$^{\S}$}

\newcommand{\myparagraph}[1]{\vspace{1ex}\noindent\textbf{#1.}\hspace{1em}}

\newcommand{\myurl}[1]{\url{#1}}

\newenvironment{myexample}{\smallskip\begin{example}\em}{\end{example}\smallskip}

\newcommand{\our}{\textsf{CEA}\xspace}
\newcommand{\rsn}{\textsf{RSNs}\xspace}
\newcommand{\mtranse}{\textsf{MTransE}\xspace}
\newcommand{\iptranse}{\textsf{IPTransE}\xspace}
\newcommand{\jape}{\textsf{JAPE}\xspace}
\newcommand{\bootea}{\textsf{BootEA}\xspace}
\newcommand{\gcn}{\textsf{GCN-Align}\xspace}

\newcommand{\gm}{\textsf{GM-Align}\xspace}

\newcommand{\mc}{\textsf{MuGNN}\xspace}
\newcommand{\na}{\textsf{NAEA}\xspace}
\newcommand{\rd}{\textsf{RDGCN}\xspace}
\newcommand{\hgcn}{\textsf{HGCN}\xspace}
\newcommand{\dbps}{\texttt{DBP15K}\xspace}
\newcommand{\dbpsz}{$\texttt{DBP15K}_\texttt{ZH-EN}$\xspace}

\newcommand{\srp}{\texttt{SRPRS}\xspace}

\newcommand{\sota}{state-of-the-art\xspace}

\let\vec\mathbf
\long\def\symbolfootnote[#1]#2{\begingroup%
\def\thefootnote{\fnsymbol{footnote}}\footnotetext[#1]{#2}\endgroup}

\def\BibTeX{{\rm B\kern-.05em{\sc i\kern-.025em b}\kern-.08em
    T\kern-.1667em\lower.7ex\hbox{E}\kern-.125emX}}
\begin{document}

\title{Collective Entity Alignment via Adaptive Features
}

\author{%
{Weixin Zeng\authorone{} \authorsep{}
 Xiang Zhao \authorcop{}\authorone{} \authorthree{} \authorsep{}
 Jiuyang Tang\authorone{} \authorthree{} \authorsep{}
 Xuemin Lin\authortwo{} \authorsep{}
}\\%
\vspace{0.5mm}\\
{\fontsize{10}{10}\selectfont\itshape{\authorone{}Science and Technology on Information Systems Engineering Laboratory, National University of Defense Technology, China}}\\%
{\fontsize{10}{10}\selectfont\itshape{\authortwo{}The University of New South Wales, Australia}}\\%
{\fontsize{10}{10}\selectfont\itshape{\authorthree{}Collaborative Innovation Center of Geospatial Technology, China}}\\%
}


\maketitle

\symbolfootnote[1]{Corresponding Author: Xiang Zhao (xiangzhao@nudt.edu.cn)}

\begin{abstract}
Entity alignment (EA) identifies entities that refer to the same real-world object but locate in different knowledge graphs (KGs), and has been harnessed for KG construction and integration. When generating EA results, current solutions treat entities independently and fail to take into account the interdependence between entities.
To fill this gap, we propose a collective EA framework. We first employ three representative features, i.e., structural, semantic and string signals, which are adapted to capture different aspects of the similarity between entities in heterogeneous KGs.
In order to make collective EA decisions, we formulate EA as the classical stable matching problem, which is further effectively solved by deferred acceptance algorithm. Our proposal is evaluated on both cross-lingual and mono-lingual EA benchmarks against state-of-the-art solutions, and the empirical results verify its effectiveness and superiority.
\end{abstract}


\section{Introduction}
Over recent years, a large number of knowledge graphs (KGs) have been constructed, whereas none of them can reach \emph{full coverage}.
These KGs, however, usually contain complementary contents, making it compelling to study the integration of heterogeneous KGs.
To incorporate the knowledge from \emph{target} KGs into the \emph{source} KG, an indispensable step is entity alignment (EA).
For each entity in the source KG, EA aims to discover its equivalent entities in target KGs.

Most state-of-the-art EA strategies assume that equivalent entities in different KGs have  similar neighbouring information. Consequently, they harness representation learning frameworks, e.g., TransE~\cite{DBLP:conf/ijcai/ChenTYZ17}, graph convolutional network (GCN)~\cite{DBLP:conf/emnlp/WangLLZ18}, recurrent skipping networks (RSNs)~\cite{DBLP:conf/icml/GuoSH19}, to model structural feature.
When generating EA results for entities in test set, these embedding-based solutions handle source entities separately and return a list of ranked target entities for each source entity with confidence (similarity) scores. The top ranked target entity is aligned to source entity.

Compared with conventional methods~\cite{DBLP:conf/cikm/GalarragaPS13}, embedding-based EA methods require less human involvement in feature engineering and can be easily scaled to large KGs.
Additionally, unlike typical entity resolution (ER) approaches~\cite{DBLP:conf/sigmod/DasCDNKDARP17,DBLP:conf/sigmod/MudgalLRDPKDAR18} that operate on \emph{relational tables}, EA centers on \emph{graph-structured data}, i.e., KGs.

State-of-the-art solutions treat entities independently when determining EA results.
However, there is often additional interdependence between different EA decisions, i.e., a target entity is less likely to be the match to a source entity if it is aligned to another source entity with higher confidence.
Sufficiently modelling such \emph{collective} signals would reduce mismatches and lead to higher aligning accuracy.
Noteworthy is that here we highlight the overlook of \emph{collective} signals during \emph{decision making} process, while we acknowledge the use of structural (collective) information as a useful feature during feature generation process.

In order to address the shortages of current EA solutions, we establish \our, a collective entity alignment framework.
\our first exploits structural, semantic and string-level features to capture different aspects of the similarity between entities in source and target KGs. These features are representative and generally available, which are effectively modelled and adapted by our proposed feature generation strategies.
Then to capture the interdependence between EA decisions, we formulate EA as the classic stable matching (marriage) problem (SMP)~\cite{gale1962college}, with entity preference lists constructed by using these features.
The problem is further addressed by deferred acceptance algorithm (DAA)~\cite{DBLP:journals/ijgt/Roth08} with high effectiveness and efficiency.
We empirically evaluate \our on both cross-lingual and mono-lingual EA tasks against 11 state-of-the-art methods, and the comparative results demonstrate its superiority. 

\section{Problem Definition and Related Work}
\label{ref}
\myparagraph{Problem definition}
Given two KGs, source KG $G_1 = (E_1, R_1, T_1)$ and target KG $G_2 = (E_2, R_2, T_2)$, where $E$, $R$ and $T$ represent entities, relations and triples, respectively, denote the seed entity pairs as $S = \{(u,v)|u\in E_1, v\in E_2, u \leftrightarrow v\}$, where $\leftrightarrow$ represents equivalence, EA task can be defined as discovering equivalent target entities for sources entities in the test set.

\myparagraph{EA using structural information}
Relying on \emph{structural information}, most efforts on EA harness KG embedding for aligning due to its simplicity, generality, and ability of dealing with large-scale data.
Initially, they utilize KG representation methods to encode structural information and embed KGs into individual low-dimensional spaces. Then the embedding spaces are aligned using seed entity pairs.
Some methods~\cite{DBLP:conf/semweb/SunHL17,DBLP:conf/icml/GuoSH19} directly project entities in different KGs into the same embedding space by fusing the training corpus first.
In accordance to distance in unified embedding space, equivalent elements in different KGs can be aligned.

\myparagraph{EA using multiple sources of information}
In addition to structural information, existing research also investigates to incorporate \emph{attribute information}, e.g., attribute types~\cite{DBLP:conf/semweb/SunHL17,DBLP:conf/emnlp/WangLLZ18}, and attribute values~\cite{DBLP:conf/aaai/TrisedyaQZ19}.
Besides entity attributes and description, 
some propose to integrate the more generally available information of \emph{entity names}~\cite{ACL19,DBLP:conf/ijcai/WuLF0Y019,IJCAI19} to provide multiple views. 
Note that these methods unify multiple views of entities at \emph{representation-level}; 
in contrast, \our 
aggregates features at \emph{outcome-level}. 

\section{Feature Generation}
\label{fg}
Three generally available features are adapted to tackle EA.

\myparagraph{Structural information}
In this work, following~\cite{DBLP:conf/emnlp/WangLLZ18}, we harness GCN~\cite{DBLP:journals/corr/KipfW16} to encode the neighbourhood information of entities as real-valued vectors. 
In the interest of space, we leave out the implementation details, which can be found in~\cite{DBLP:conf/emnlp/WangLLZ18}. 

Given the learned structural embedding matrix $\vec Z$, the similarity between two entities can be calculated by cosine similarity. We denote the structural similarity matrix as $\vec {M^s}$, where rows represent source entities, columns denote target entities and each element in the matrix denote structural similarity score between a pair of source and target entities.


\myparagraph{Semantic information}
Entity name can be exploited both from the semantic and string similarity level.  
We choose averaged word embeddings to capture the semantic meaning on account of its simplicity and generality. It does not require special training corpus, and can represent semantics in a concise and straightforward manner. 
For a KG, the name embeddings of all entities can be denoted in matrix form as $\vec N$ and cosine similarity is harnessed to capture entity proximity. We denote semantic similarity matrix as $\vec {M^n}$. 


\myparagraph{String information}
Current methods mainly capture semantic information of entities, as semantics can be easily encoded as embeddings, facilitating the fusion of different feature representations~\cite{IJCAI19,ACL19}. Plus, semantic feature can also work well in cross-lingual scenarios by referring to external sources, e.g., pre-trained multilingual word embeddings.

In our work, we contend that string information, which has been largely overlooked by existing embedding-based EA literature, is also beneficial, since:
\begin{inparaenum} [(1)]
	\item string similarity is especially useful in tackling mono-lingual EA task and cross-lingual EA task where KG pair is very close (e.g., English and German); and
	\item string similarity does not rely on external resources and is not affected by out-of-vocabulary problem which tends to restrain the effectiveness of semantic information. 
\end{inparaenum} 
In particular, we adopt Levenshtein distance, a string metric for measuring the difference between two sequences. 
We denote the corresponding string similarity matrix as $\vec {M^l}$. 

\section{Collective Entity Alignment}
\label{ff}
\myparagraph{EA via stable matching}
After obtaining feature-specific similarity matrices, we combine them with equal weights and generate the fused matrix $\vec M$. Then EA results can be determined in an independent fashion that has been adopted by state-of-the-art methods.
Specifically, for each source entity $u$, we retrieve its corresponding row entry in $\vec M$, and rank the elements in a descending order.
The top ranked target entity is considered as the match.

Nevertheless, this way of generating EA pairs fails to consider the interdependence between different EA decisions. 
To adequately model such coherence, we formulate EA as the stable matching problem~\cite{gale1962college}.
Concretely, it is proved that for any two sets of members with the same size, each of whom provides a ranking of the members in the opposing set, there exists a bijection of the two sets such that no pair of two members from the opposite side would prefer to be matched to each other rather than their assigned partners~\cite{DBLP:conf/ccs/DoernerES16}. This set of pairing is also called \emph{a stable matching}.
As input to the stable matching process, the \emph{preference list} of an entity is defined as the set of entities in opposing side ranked by similarity scores in a descending order.


\begin{figure}[h]
	\centering
	\includegraphics[width=0.8\linewidth]{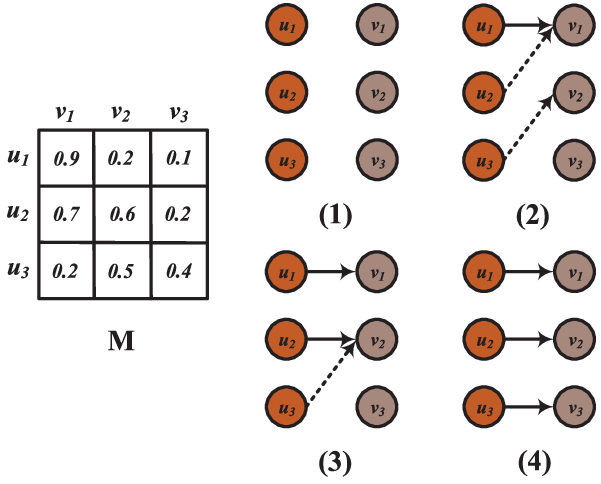}
	\caption{EA as SMP using DAA.}
	\label{fig:daa}%
\end{figure}

\myparagraph{Algorithmic solution}
To solve SMP, we adopt deferred acceptance algorithm (DAA)~\cite{DBLP:journals/ijgt/Roth08}, which works as follows:
\begin{enumerate} [(1)]
	\item In the first round, each source entity proposes to the target entity it prefers most, and then
	the target entity is provisionally matched to the source entity it most prefers so far, and that source entity is likewise provisionally matched to the target entity.
	
	\item In each subsequent round, each \emph{unmatched} source entity proposes to the most-preferred target entity to which it has not yet proposed (regardless of whether the target entity is already matched), and then each target entity accepts the proposal if it is currently not matched or if it prefers the new proposer over its current match (in this case, it rejects its provisional match and becomes unmatched).
	
	\item The process of (2) is repeated until every source entity is settled, i.e., find its match.
\end{enumerate}

\begin{myexample}
	The example in Figure~\ref{fig:daa} illustrates the algorithm.
	(1) Suppose there are three source entities $u_1, u_2, u_3$ and three target entities $v_1, v_2, v_3$;
	(2) Given the fused similarity matrix $\vec M$, where the values in an entity's corresponding row/column denote its preferences (larger value is preferred). Source entities first make proposals to target entities. Both $u_1$ and $u_2$ propose to $v_1$ and $u_3$ proposes to $v_2$. Since $v_1$ prefers $u_1$ to $u_2$, it accepts $u_1$ and rejects $u_2$, and $v_2$ accepts its only proposer $u_3$;
	(3) In the next round, $u_2$ is the only source entity that is unmatched and it proposes to its second most preferable entity $v_2$. Although $v_2$ is temporally matched to $u_3$, it accepts $u_2$'s offer and rejects $u_3$ since it prefers $u_2$ to $u_3$;
	(4) In the final round, the only unmatched source entity $u_3$ proposes to $v_3$ and they form a match.
\end{myexample}

This algorithm guarantees that every entity gets matched. At the end, there cannot be a source entity and a target entity both unmatched, as this source entity must have proposed to this target entity at some point (since a source entity will eventually propose to every target entity, if necessary) and, being proposed to, the target entity would necessarily be matched (to a source entity) thereafter. Also, the ``marriages'' are stable. Let $u$ and $v$ both be matched, but not to each other. Upon completion of the algorithm, it is not possible for both $u$ and $v$ to prefer each other over their current partners.

\myparagraph{Discussion}
Theoretically, EA can also be formed as the maximum weighted bipartite matching problem that requires the elements from two sets to be paired one-to-one \emph{and} to have the largest sum of pairwise utilities (similarity scores).
In this way, the classic Hungarian algorithm may be applied, which however could be computationally prohibitive for EA ($O(n^4)$).
Nevertheless, stable matching is more desirable, since it produces a matching result where no participants have incentives to deviate from and requires much less time. 

\section{Experiments}
\label{exp}

\myparagraph{Experimental setting}
Following previous works, we adopt \dbps~\cite{DBLP:conf/semweb/SunHL17} and \srp~\cite{DBLP:conf/icml/GuoSH19} as evaluation datasets.
Regarding parameter setting, in {\itshape entity name representation}, we utilize the fastText embedding as word embedding and the multilingual word embeddings are obtained from MUSE\footnote{\url{https://github.com/facebookresearch/MUSE}}. The source code of \our is publicly available\footnote{\url{https://github.com/DexterZeng/CEA}}.

We utilize the accuracy of alignment results as evaluation metric. It is defined as the number of correctly aligned source entities divided by the total number of source entities.
11 state-of-the-art EA methods are adopted for comparison, which can be divided into two groups, methods merely using structural information and methods using information external to structural information.

\begin{table}[htbp]
	\centering
	\begin{threeparttable}
		\caption{Results on \dbps.}
		\begin{tabular}{cccc}
			\toprule
			& ZH-EN & JA-EN & FR-EN \\
			\midrule
			\mtranse$^\star$~\cite{DBLP:conf/ijcai/ChenTYZ17} & 0.308 & 0.279 & 0.244 \\
			\iptranse$^\star$~\cite{DBLP:conf/ijcai/ZhuXLS17} & 0.406 & 0.367 & 0.333 \\
			\bootea~\cite{DBLP:conf/ijcai/SunHZQ18} & 0.629 & 0.622 & 0.653 \\
			\rsn$^\circ$~\cite{DBLP:conf/icml/GuoSH19}  & 0.581 & 0.563 & 0.607 \\
			\mc~\cite{DBLP:conf/acl/CaoLLLLC19} & 0.494 & 0.501 & 0.495 \\
			\na~\cite{DBLP:conf/ijcai/ZhuZ0TG19}  & 0.650 & 0.641 & 0.673 \\
			\midrule
			\gcn~\cite{DBLP:conf/emnlp/WangLLZ18} & 0.413 & 0.399 & 0.373 \\
			\jape~\cite{DBLP:conf/semweb/SunHL17}  & 0.412 & 0.363 & 0.324 \\
			\rd~\cite{DBLP:conf/semweb/SunHL17} & 0.708 & 0.767 & 0.886 \\
			\hgcn~\cite{wu2019jointly}  &0.720 & 0.766 & 0.892 \\
			\gm~\cite{ACL19} & 0.679 & 0.740 & 0.894 \\
			\our   & \textbf{0.787} & \textbf{0.863} & \textbf{0.972} \\
			\bottomrule
		\end{tabular}%
		\label{tab:dbp}%
		\begin{tablenotes}
			\footnotesize {
				\item[1] The results of $\star$-marked methods are obtained from~\cite{DBLP:conf/ijcai/SunHZQ18}. We run the source codes of $\circ$-marked methods to attain the results. The rest are from their original papers.
			}
		\end{tablenotes}
	\end{threeparttable}
\end{table}%

\myparagraph{Evaluation results}
From Tables~\ref{tab:dbp} and~\ref{tab:srp}, it reads that our model \our consistently outperforms all baselines.
The superiority can be attributed to the fact that: (1) three representative sources of information, i.e, structural, semantic and string-level features, are leveraged to offer more comprehensive signals for EA; and (2) source entities are aligned collectively, which can avoid frequently appearing situations where multiple source entities are aligned to the same target entity.

In specific, solutions in the first group merely harness structural information for aligning.
\mtranse obtains the worst results.
\iptranse achieves better performance than \mtranse as it adopts relational path for learning structural embedding, and utilizes an iterative framework to augment training set.
Additionally, \rsn enhances the results by taking into account of long-term relational dependencies between entities, which can capture more structural signals for alignment, while \bootea devises a carefully designed alignment-oriented KG embedding framework, with one-to-one constrained bootstrapping strategy.

On \dbps, \mc also outperforms \iptranse since it utilizes a multi-channel graph neural network that captures different levels of structural information. Nevertheless, its results are still inferior to \bootea and \rsn. \na attains the highest results within this group, as its neighbourhood-aware attentional representation method can make better use of KG structures and learn more comprehensive structural representation.
Nevertheless, we implement their source codes on \srp and the corresponding results are much worse, as shown in Table~\ref{tab:srp}.

\begin{table}[htbp]
	\centering
	\begin{threeparttable}
		\caption{Results on \srp.}
		\begin{tabular}{ccccc}
			\toprule
			& EN-FR & EN-DE & DBP-WD & DBP-YG \\
			\midrule
			\mtranse$^\star$ & 0.251 & 0.312 & 0.223 & 0.246 \\
			\iptranse$^\star$ & 0.255 & 0.313 & 0.231 & 0.227 \\
			\bootea$^\star$ & 0.313 & 0.442 & 0.323 & 0.313 \\
			\rsn$^\star$  & 0.347 & 0.487 & 0.388 & 0.400 \\
			\mc & 0.131 & 0.245 & 0.151 & 0.175 \\
			\na  & 0.195 & 0.321 & 0.215 & 0.211 \\
			\midrule
			\gcn$^\star$ & 0.155 & 0.253 & 0.177 & 0.193 \\
			\jape$^\star$  & 0.256 & 0.320 & 0.219 & 0.233 \\
			\rd & 0.672 & 0.779 & 0.827 & 0.846 \\
			\hgcn  & 0.670 & 0.763 & 0.823 & 0.822 \\
			\gm & 0.627 & 0.677 & 0.815 & 0.827 \\
			\our   & \textbf{0.962} & \textbf{0.971} & \textbf{0.998} & \textbf{0.999} \\
			\bottomrule
		\end{tabular}%
		\label{tab:srp}%
		\begin{tablenotes}
			\footnotesize {
				\item[1] The results of $\star$-marked methods are obtained from~\cite{DBLP:conf/icml/GuoSH19}. We run the source codes of the rest methods to attain the results.
			}
		\end{tablenotes}
	\end{threeparttable}
\end{table}%

Noteworthily, the overall performance on \srp are worse than \dbps, as the KGs in \dbps are much denser than those in \srp and contain more structural information~\cite{DBLP:conf/icml/GuoSH19}.
On \srp, where KGs are with real-life degree distributions, \rsn achieves the best results since the long-term relational dependencies it captures are less sensitive to entity degrees. This is verified by the fact that the results of \rsn exceed \bootea, in contrast to results on \dbps.

Within the second group, taking advantage of attribute information, \gcn and \jape outperform \mtranse on \dbps, whereas on \srp, \gcn achieves worse results than \mtranse and \jape attains similar accuracy scores to \mtranse. This reveals that attribute information is quite noisy and might not guarantee consistent performance.
\rd, \hgcn and \gm outperform all the other methods except for \our, as they harness entity name information as the inputs to their overall framework for learning entity embeddings, and the final entity representation encodes both structural and semantic signals, providing a more comprehensive view for alignment.
Nevertheless, \our outperforms these name-based methods by a large margin, since we aggregate features on outcome-level instead of representation-level.


Notably, on mono-lingual datasets, \our advances the accuracy to almost 1.000. This is because entity names in DBpedia, YAGO and Wikidata are nearly identical, where string-level feature is extremely effective.
In contrast, although semantic information is also useful, not all words in entity names can find corresponding entries in external word embeddings, which hence limits its effectiveness.
The fact that a simple string-level feature can achieve ground truth results on current benchmarks also encourages us to build more challenging mono-lingual EA datasets, which is left for future work.

\begin{table}[htbp]
	\centering
	\caption{Ablation results on \srp and \dbpsz}
	\begin{tabular}{cccccc}
		\toprule
		& EN-FR & EN-DE & DBP-WD & DBP-YG& ZH-EN \\
		\midrule
		\our   & 0.962 & 0.971 & 0.998 & 0.999 & 0.787 \\
		\textsf{w/o C}  & 0.914 & 0.925 & 0.986 & 0.994 & 0.701 \\
		\textsf{w/o $\vec {M^s}$} & 0.915 & 0.968 & 0.998 & 0.998 & 0.622 \\
		\textsf{w/o $\vec {M^n}$} & 0.947 & 0.967 & 0.998 & 0.998 & 0.507 \\
		\textsf{w/o $\vec {M^l}$} & 0.782 & 0.863 & 0.915 & 0.937 & 0.778 \\
		\bottomrule
	\end{tabular}%
	\label{tab:ablation}%
\end{table}%

\myparagraph{Ablation study}
We perform ablation study to gain insights into the components of \our. The results on \srp and \dbpsz are presented in Table~\ref{tab:ablation}

We first examine the contribution of the collective scheme.
As can be observed from Table~\ref{tab:ablation}, without collective EA, the performance drops on all datasets, revealing the significance of considering the interdependence between EA decisions.

We then test the feasibility of our proposed features.
On cross-lingual datasets, removing structural information consistently brings performance drop, showing its stable effectiveness across all language pairs.
In comparison, semantic information plays a more important role on distantly-related language pairs, e.g., Chinese--English, whereas string-level feature is significant for aligning closely-related language pairs, e.g., English--French.
On mono-lingual datasets, removing structural or semantic information does not hurt the accuracy, while pruning string-level feature results in a considerable accuracy drop.
This unveils that string-level feature is quite useful on datasets where entity names are similar. 

\section{Conclusion}
In this paper, we have investigated the problem of EA for heterogeneous knowledge fusion. Existing EA methods overlook the opportunity to boost performance by constructing a collective alignment solution. In response, we propose to incorporate multiple adaptive features and resort to stable matching for collective EA. The resultant method is experimentally verified to be superior to \sota options.

%
\vspace{2.5ex}
\myparagraph{Acknowledgment}
This work was partially supported by NSFC under grants Nos. 61872446, 61902417, and 71971212, and NSF of Hunan province under grant No. 2019JJ20024.

{
\bibliographystyle{abbrv}
\bibliography{icde20}
}

\end{document}